\documentclass{article}
\usepackage[utf8]{inputenc}
\usepackage{authblk}
\usepackage{booktabs}
\usepackage{graphicx}

\title{Machine learning methods for accurate delineation of tumors in PET images}

\author[1]{Jakub Czakon}
\author[1]{Filip Drapejkowski}
\author[1]{Grzegorz Żurek}
\author[1]{\\Piotr Giedziun}
\author[2]{Jacek Żebrowski}
\author[1]{Witold Dyrka\thanks{Corresponding author: witold.dyrka@stermedia.pl}}
\affil[1]{Stermedia Sp. z o. o., CancerCenter.eu team\thanks{Website: CancerCenter.eu}, ul.\,A.\,Ostrowskiego\,13, Wrocław, Poland}
\affil[2]{Lower Silesian Oncology Center, Department of Nuclear Medicine - PET-CT Laboratory, pl. L. Hirszfelda 12, Wrocław, Poland}

\date{}

\begin{document}

\maketitle

\begin{abstract}

In oncology, Positron Emission Tomography imaging is widely used in diagnostics of cancer metastases, in monitoring of progress in course of the cancer treatment, and in planning radiotherapeutic interventions. Accurate and reproducible delineation of the tumor in the Positron Emission Tomography scans remains a difficult task, despite being crucial for delivering appropriate radiation dose, minimizing adverse side-effects of the therapy, and reliable evaluation of treatment. In this piece of research we attempt to solve the problem of automated delineation of the tumor using 3d implementations of the spatial distance weighted fuzzy c-means, the deep convolutional neural network and a dictionary model. The methods, in diverse ways, combine intensity and spatial information.

\end{abstract}

\section{Motivation}
Positron Emission Tomography (PET) is a nuclear imaging technique used to monitor metabolic activity of the body. The PET system detects gamma rays generated when electrons collide with positrons emitted by radiolabelled biologically active molecules, such as fluorodeoxyglucose. In oncology, PET is widely used in diagnostics of cancer metastases, in monitoring of progress in course of the cancer treatment, and also in planning radioterapheutic interventions. In the last case, PET scans, often in combination with Computer Tomography (CT) scans, are used to delineate biologically active parts of tumors, which are then treated with ionizing radiation. Accurate delineation of the tumor is crucial for delivering appropriate radiation dose and minimizing adverse side-effects of the therapy. However, the problem is not trivial as evidenced by substantial intra- and interobserver variations when delineation is performed manually by experts\cite{nje08}. Therefore it is not surprising that automated methods for accurate and reproducible delineation are sought.

\section{Methods}

In this piece of research we evaluated 3d implementations of Spatial Distance Weighted Fuzzy C-Means (SDWFCM) of Guo et al. \cite{guo15}, Dictionary-based Model (DICT) of Dahl and Larser \cite{dah11} and Convolutional Neural Network (CNN) \cite{kri12}, and compared them to classical approaches clustering approaches: K-Means (KM) \cite{bez81} and Gaussian Mixture Models (GMM) \cite{zen14}.

In order to assure consistency of the input data, images detected to be sharper than an arbitrary threshold (after convolution with the 3d Laplacian kernel) were slightly blurred using the Gaussian filter. The step was omitted in the GMM and CNN case. In order to clean resulting labels from small artifacts, the labels were subject to subsequent morphological opening and closing, except the CNN method.

We used $sklearn$ implementations of KM and GMM methods.

\subsection{Spatial Distance Weighted Fuzzy C-Means}
The method is a spatial information-aware extension of the Fuzzy C-Means clustering algorithm (FCM) \cite{bez81}. Both methods assign to each point a coefficient of belonging to one of $c$ clusters. Similarly to FCM, SDWFCM starts by calculating cluster centers and Euclidean distances based on regular features (such as image intensity). However, in the next step the distance matrix is modified with weight based on spatial distance of each pixel to the cluster center. Then, coefficients of belonging are calculated based on the modified distance matrix. The procedure repeats until convergence or maximum runtime. Finally the tumor label is assigned to $f$ most intense clusters. Parameters for the SDWFCM algorithm (as well as for KM and GMM) were optimized for the training set (see Tabl.~\ref{results} and ref.~\cite{guo15}).

\subsection{Dictionary-based Model}
\label{dict}
The general idea of the method is to assign each patch of the test image, the label which corresponds to the most similar patch in the dictionary. The dictionary, which consists of a set of patch-label pairs is built in two steps. First, a sample of training patches is used to seed the dictionary. Based on the label similarity threshold, similar patches are clustered and averaged. Then, a version of the vector quantization technique is used to adjust the dictionary patches such that real labels of its matches are as close to binarized dictionary labels (also adjusted in the training). Finally, the labeling algorithm assigns binary labels of the most similar dictionary patch. In our implementation, the labeling window walks pixel by pixel and each pixel gets the label of the best fitting patch or an average label, which is then binarized using the final labeling threshold $th$. DICT was set up with the following parameters: 3x3x3 patches were utilized, 10\% of the training patches were used to seed the dictionary using similarity threshold 0.5; all training patches were used for adjusting the dictionary in 10 iterations with the step coefficient $\tau$ 0.05 (see ref.~\cite{dah11}).

\subsection{Convolutional Neural Network}
\label{cnn}
Deep convolutional neural networks have proved to be one of the most efficient tools in image processing.  For the tumor delineation problem we tested many CNN architectures consisting of up to dozen layers, including three or four convolutional layers. The input of the network was a batch of 2d or 3d patches from the PET tumor images. The network was trained using the AdaGrad stochastic gradient descent algorithm \cite{duc11}, which aimed at minimizing the Euclidean distance between the output of the final layer of the network and the ground truth. To account for a relatively small number of training samples, the set of available data was artificially augmented with rotationally transformed samples. The best architecture, selected in the 5-fold cross-validation process, was the 3D convolutional neural network with 3 convolutional layers followed by RELU activations with ascending number of filters (Fig.~1).

\begin{figure}[!ht]
  \label{figcnn}
  \caption{\textbf{CNN architecture.} The network input are 3d patches (5x10x10 pixels) of the PET scan. The network output are 3d patches (the same dimensions as input) with predicted binary tumor segmentation. Neighbouring output patches are combined to give a binary image of the whole scan (not shown). Notations: CONV - convolutional layer, DROP - drop-out layer with ratio $r$, ACTIV $fun$ - activation layer with function $fun$.}
  \centering
  \includegraphics[width=4.5in]{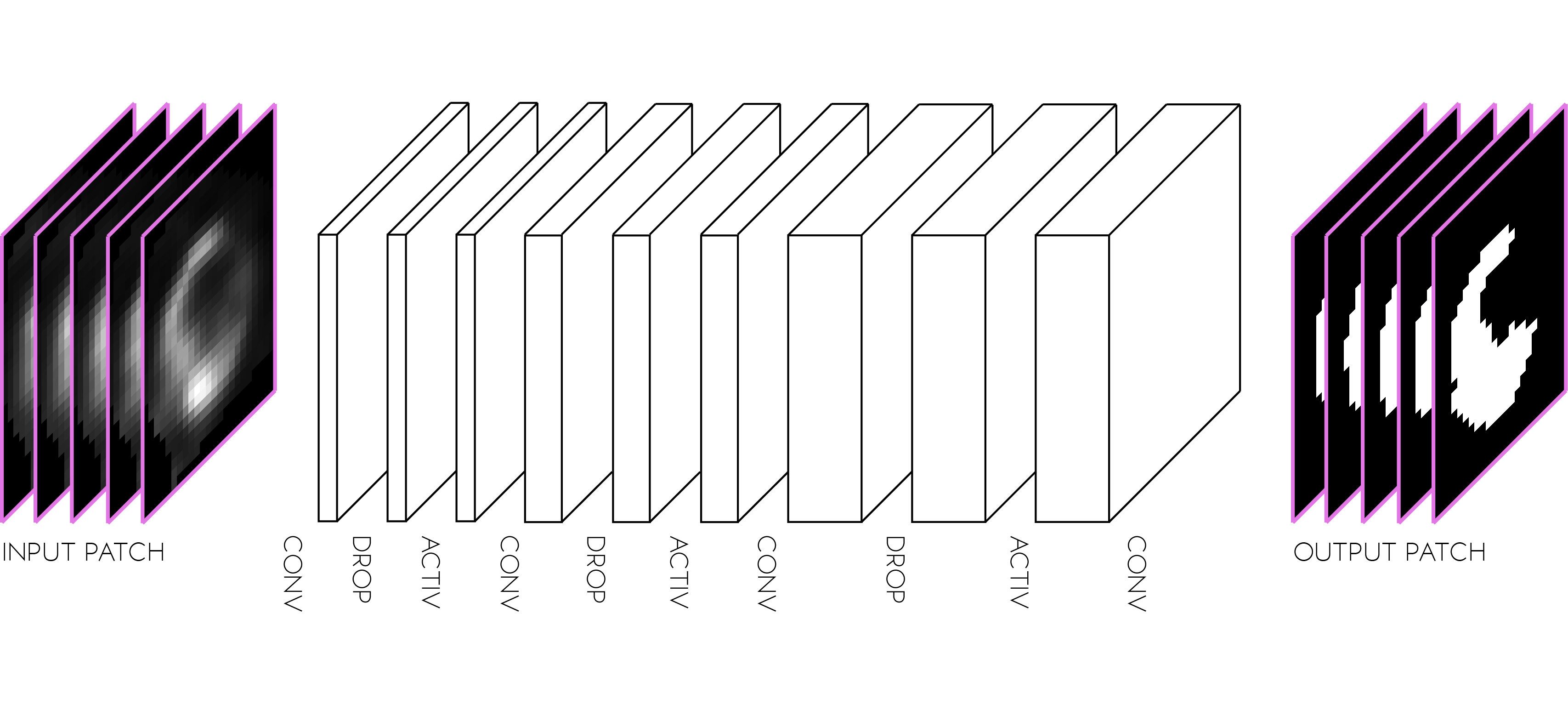}
\end{figure}

\section{Results}

The training dataset provided by organizers of the PETSEG challenge consisted of 4 scans from 3 clinical samples, 9 scans from 3 phantoms and 6 simulated scans in 2 sets. We assessed efficiency of the methods on the training set in terms of the Dice coefficient. As the scores were calculated on the training set, they are rather upper estimates of the real performance of the methods.

\begin{table}[h!]
\centering
\caption{\textbf{Dice coefficients for training data.} Notations: KM - K-Means, GMM - Gaussian Mixture Model, SDWFCM - Spatial Distance Weighted Fuzzy C-Means, DICT - Dictionary-based model, CNN - Convolutional Neural Network; $k$/$n$/$c$ - number of clusters, $f$ - number of selected most intense clusters, $m$ - degree of fuzzy classification, $\lambda$ - weight of spatial features, $nb$ - number of neighbours.}
\label{results}
\begin{tabular}{@{}llrrrrrr@{}}
\toprule
method & parameters                 & \multicolumn{1}{l}{avg} & \multicolumn{1}{l}{med.} & \multicolumn{1}{l}{clin.} & \multicolumn{1}{l}{phan.} & \multicolumn{1}{l}{simul.} & \multicolumn{1}{l}{balanced} \\ \midrule
KM     & k=2, f=1                   & 0.82                        & 0.81                       & 0.77                         & 0.79                        & 0.90                          & 0.82                         \\
GMM    & n=4, f=1                   & 0.83                        & 0.83                       & 0.77                         & 0.80                        & 0.90                          & 0.82                         \\
SDWFCM    & c=2, f=1\\ 
       & m=2, $\lambda$=0.5\\
       & nb=1 & 0.82                        & 0.82                       & 0.76                         & 0.81                        & 0.88                          & 0.81                         \\
DICT   & see Sec.~\ref{dict}   & 0.82                        & 0.81                       & 0.77                         & 0.78                        & 0.90                          & 0.82                         \\
CNN    & see Sec.~\ref{cnn}                    & \textbf{0.86}               & \textbf{0.89}              & \textbf{0.78}                & \textbf{0.86}               & \textbf{0.91}                 & \textbf{0.85}                \\
\bottomrule
\end{tabular}
\end{table}

\noindent
The best results were obtained using CNN (Dice score mean: 0.86, median: 0.89). The remaining methods performed slightly worse than CNN and similarly to each other (Dice score mean: 0.82-0.83, median: 0.81-0.83). In general, simulated data proved to be the easiest (mean Dice score 0.88-0.91), while clinical data were the hardest (mean Dice score 0.76-0.78), regardless the method. A single most difficult case was \textit{clinical\_1} for which none of the algorithms exceeded Dice score of 0.69. It is of note, that the Gaussian blurring had a positive impact on segmentation of this case. In fact \textit{clinical\_1, clinical\_3} and \textit{phantom\_2.1} were the only cases where CNN did not reach Dice score of 0.80. Interestingly, CNN was always better than other methods for phantom data, with average margin over the second best result of almost 7\%.

Looking more into details, CNN and DICT achieved the highest sensitivity (0.89 and 0.85, respectively), while SDWFCM achieved the highest precision (0.85). In accordance with these results, CNN and DICT were more prone to overestimate the total volume of the tumor (on average by 10\%), while SDWFCM lined towards underestimated volumes (on average by 3\%). All methods overestimated the volume of \textit{clinical\_1} (by 30-49\%), \textit{phantom\_2} (9-58\%) and \textit{phantom\_3} (8-47\%), while the volume of \textit{clinical\_3} was generally underestimated (12-26\%). Interestingly, while CNN was the method which overestimated the volume of clinical samples the most (22\%), its volume predictions were closest to the ground truth for the phantoms (+9\%). Accordingly, the contour mean distance of CNN predictions was the worst for clinical samples (2.28 in comparison to 1.81 by DICT), and the best for phantom samples (0.72 in comparison to 1.30 by DICT). 

Overally, our early results suggest that deep convolutional neural networks can enable a step forward also in the PET image segmentation. This claim should be further validated with more testing, especially on clinical samples, for which performance gain using CNN was less impressive than for phantom data.

\paragraph{Acknowledgements.}
This research has been conducted under the Support Programme of the Partnership between Higher Education and Science and Business Activity Sector financed by City of Wroclaw. The authors thank Jakub Szewczyk for his help with preparing the manuscript, and Piotr Krajewski for continuous support of the project.

\end{document}